# A very preliminary analysis of DALL-E 2

Gary Marcus, New York University, gary.marcus@nyu.edu

Ernest Davis, New York University, davise@cs.nyu.edu

Scott Aaronson, University of Texas at Austin, scott@scottaaronson.com

**Abstract:** The DALL-E 2 system generates original synthetic images corresponding to an input text as caption. We report here on the outcome of fourteen tests of this system designed to assess its common sense, reasoning and ability to understand complex texts. All of our prompts were intentionally much more challenging than the typical ones that have been showcased in recent weeks. Nevertheless, for 5 out of the 14 prompts, at least one of the ten images fully satisfied our requests. On the other hand, on no prompt did *all* of the ten images satisfy our requests.

The DALL-E 2 system, created by OpenAI, generates original synthetic images corresponding to an input text as caption (Ramesh, Dhariwal, Nichol, Chu, & Chen, 2022)[1]. One of us was recently granted access to DALL-E 2. We asked fourteen questions, designed to assess common sense, reasoning, and its capacity to assess complex utterances. We give the full results below.

Whether results of this kind should be considered as successes for the program – what is the proper measure to use in evaluating success – depends on the intended use of the program. If the goal is to generate candidate images that a graphic artist will choose from, or choose from and edit, then the system can reasonably be measured in terms of the quality of the best result out of ten or out of one hundred To the extent that the goal is to develop artificial intelligence that can be trusted in safety-critical applications (Marcus & Davis, 2019), a much higher standard must be applied.

While our investigation was only preliminary, some tentative conclusions can be drawn:

• The visual quality of the images is stunning. We were particularly impressed with the ability to capture the top-down perspective we requested in Example 6. A commercial artist might have trouble getting DALL-E2 to deliver the exact results that they or their client require; an amateur looking for clipboard-like art with less strict expectations may well get something striking and close enough to what is needed with very little effort.

• DALL-E 2 is unquestionably extremely impressive in terms of image generation. The system succeeds in applying many diverse artistic styles to the specified subject with extraordinary fidelity and aptness, and capture their spirit: cartoons are light-hearted, impressionist paintings are peaceful and evocative, photographs of everyday scenes are naturalistic, noir photographs are subtly disturbing. Images in realistic styles are almost always physically plausible (we note exceptions in example 9 and 12 below); images in non-realistic styles conform to the particular norms of the style. Many of the images that have been published demonstrate DALL-E 2's remarkable ability to create striking surrealist images, such as the half-human, half-robotic face of Salvador Dali (after whom the program is named) included in (Ramesh, Dhariwal, Nichol, Chu, & Chen, 2022). (Our experiments did not include any captions that would particularly lend themselves to a surrealistic portrayal.)

---

[1] The DALL-E 2 system is called "unCLIP" in this article.

• Some aspects of the system's language abilities seem to be quite reliable. If a caption specifies only two or three objects, the system almost always shows all of them. If a caption specifies a feature of an object, then the image will generally show that feature somewhere, though, as we will discuss below, not necessarily on the correct object. The examples that have been published elsewhere demonstrate that DALL-E 2 can reliably follow stylistic instructions (we did this in only one of our experiments). In examples 7 and 14 below, the system reliably follows viewpoint specifications, even though 7 requires a non-canonical view of the scene.

• Compositionality, in the Fregean sense of meaning derived from parts, appears to be particularly problematic.

    • Results are often incomplete (examples 2, 3, 4, 7 below).

    • Relationships between entities are particularly challenging (examples 1-5, 7, 9). The failures in DALL-E 2 in correctly associating specified properties with objects and in placing objects in their correct relation in phrases like "a red cube on top of a blue cube" have been discussed in (Ramesh, Dhariwal, Nichol, Chu, & Chen, 2022). Similarly, Thrush et al. (2022) tested the visuo-linguistic compositional reasoning of current state-of-the-art vision and language models using the Winoground benchmark. They found that none of these, including CLIP (Radford, et al., 2021) which forms part of the basis for DALL-E 2, does significantly better than chance.

    • Anaphora and phrases like "in a similar posture" that require connecting parts of a sentence across a discourse may pose particular problems (example 1).

• Numbers may be poorly understood (example 6).

• Negation has been problematic for large language models (Ettinger, 2020) and our preliminary probing suggests similar problems here (example 12).

• Common sense. Two of our examples were designed to probe DALL-E 2 commonsense knowledge. It succeeded in example 9, but failed in example 10. One of the images in example 12 also shows a failure of common sense reasoning.

• The content filters need improvement; in example 5, we were unable to test an item with the phrase tug-of-war (describing the children's game), apparently because the substring "war" was flagged as a policy violation.

• Generalization is difficult to directly assess, because no details of the training set have been made available. What might or might not count as distribution shift is not obvious.

Shortly after DALL-E 2 was released, OpenAI CEO Sam Altman tweeted "AGI is gonna be wild."[2] However, it is reasonable to question whether DALL-E 2 constitutes progress toward solving the deep challenges of commonsense reasoning, comprehension, reliability, and so forth that would be needed for a truly general-purpose AI (Marcus & Davis, 2019). Our results provide a clearer picture of what remains to be done before a system like DALL-E 2 could genuinely be said to understand what it strikingly renders in images.

``Swimmer963'' (2022) reports on similar experiments, carried out by a user willing to invest significant time and effort in trying a variety of prompts to get a suitable image. They encountered the same kind of problems that we did, and report that in several cases they were unable to get DALL-E 2 to produce what they wanted, despite multiple attempts.

## Methods

DALL-E 2 was run at default settings. Ten images (the default) were generated for each caption. Examples 6 and 12 were run twice with changes to the caption because of technical issues discussed below.

The examples were specifically designed to probe what we conjectured might be weaknesses in DALL-E 2. We screened three proposed captions by running them through Google image search to see how easily images could be retrieved; as a result we rejected two proposed caption and modified another. It may be noted that the authors have no specific knowledge of the internals, the training set, or the abilities of DALL-E 2 beyond what is published in (Ramesh, Dhariwal, Nichol, Chu, & Chen, 2022) and the examples that have circulated in social media.

**Example 1:**

Caption: a red basketball with flowers on it, in front of blue one with a similar pattern

Images:

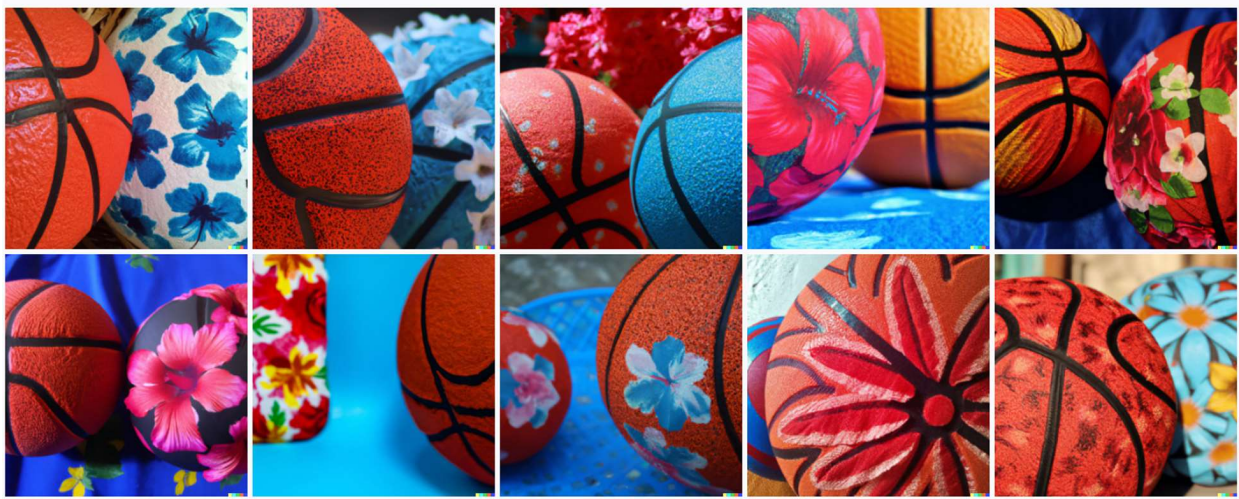

---

[2] https://twitter.com/sama/status/1511735572880011272

Discussion: Most of these have omitted some aspect of the description. Image 10 is correct. (In all these discussions, images are numbered 1-5 left to right in the top row and 6-10 left to right in the bottom row.)

**Example 2:**

Caption: a red ball on top of a blue pyramid with the pyramid behind a car that is above a toaster.

Images:

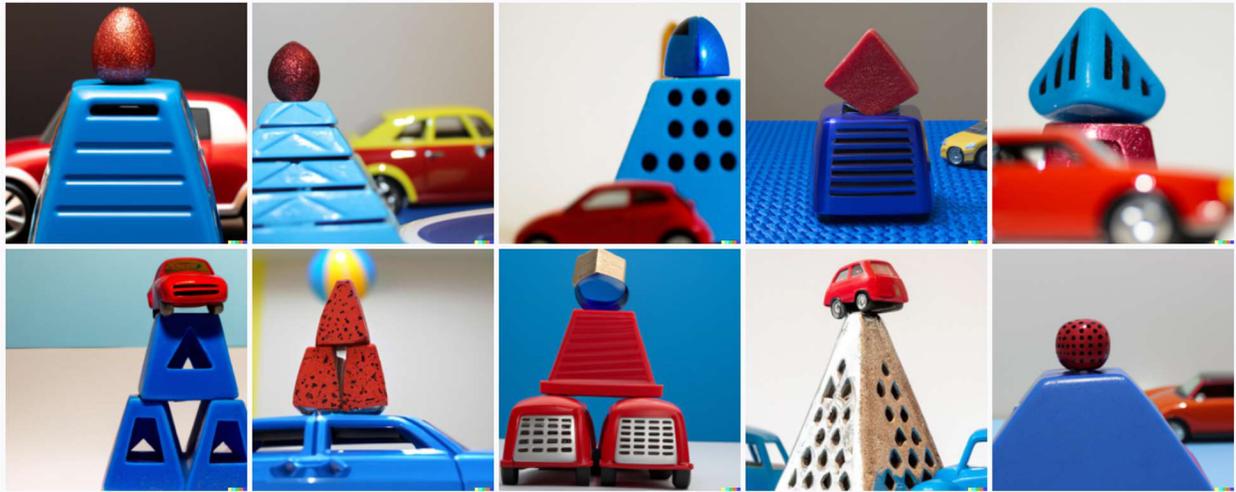

Discussion: None of these are correct. The inability of DALL-E 2 to deal properly with complex specifications is discussed in (Ramesh, Dhariwal, Nichol, Chu, & Chen, 2022).

**Example 3:**

Caption: Abraham Lincoln touches his toes while George Washington does chin-ups. Lincoln is barefoot. Washington is wearing boots.

Images:

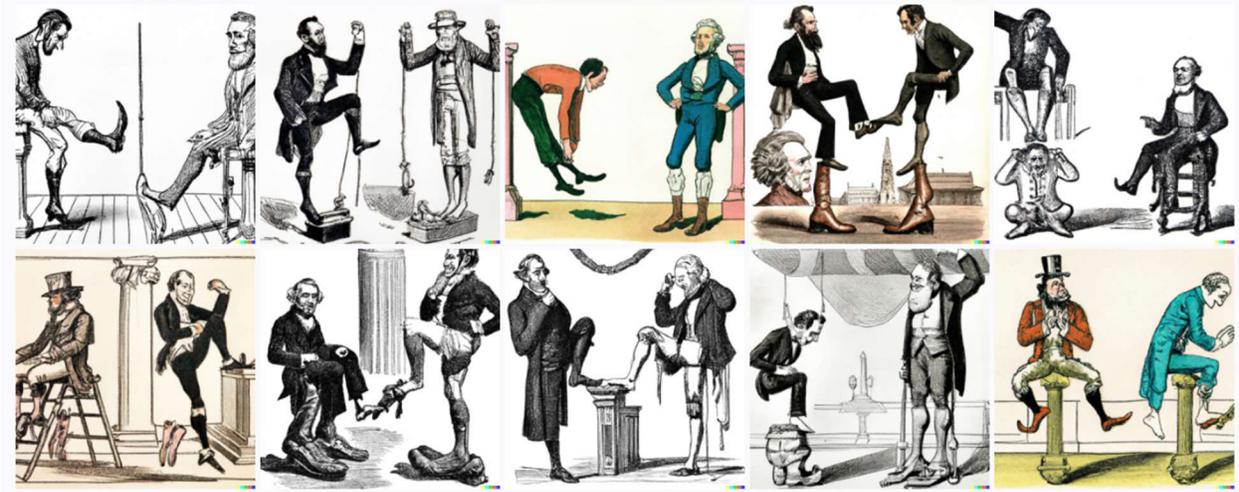

Discussion: All of the images show two men. Several feature toes and/or boots. All show the men in rather unusual postures, suggestive of exercise, but only image 3 shows (rather strangely) a man touching his toes, and none show a man doing chin-ups. Images 1 and 6 show Lincoln; none of the images seems to show Washington.

In all our examples, the ten images generated for a single caption were in a single artistic style, though that was not specified in our caption. It is particularly conspicuous in this case, because the particular style of early nineteenth-century cartoonists such as John Doyle and George Cruikshank used here is quite distinctive and unusual. Apparently, the associations between captions and images that DALL-E 2 uses (the CLIP representation (Radford, et al., 2021)) tends to associate a caption with a collection of images of one particular genre..

**Example 4:**

Caption: Supreme Court Justices play a baseball game with the FBI. The FBI is at bat, the justices are on the field.

Images:

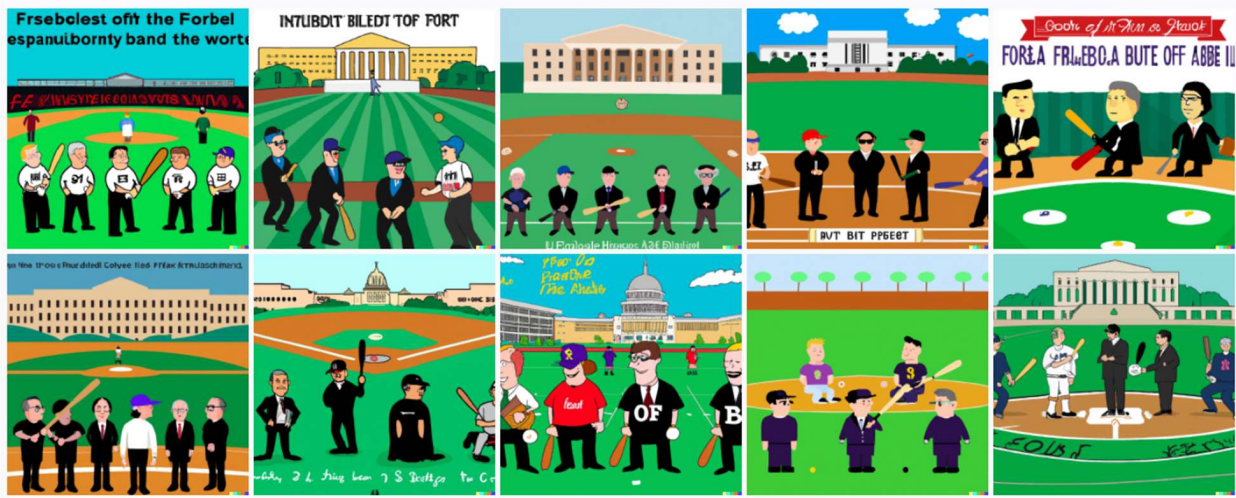

Discussion: The images mostly show people, who could well be government officials, some carrying bats, many near recognizable government buildings. But none comes very close to showing the specified situation. The test captions that have been inserted are all gibberish; this failing of DALL-E 2 is noted in (Ramesh, Dhariwal, Nichol, Chu, & Chen, 2022).

**Example 5:**

Caption: A donkey is playing tug-of-war against an octopus. The donkey holds the rope in its mouth. A cat is jumping over the rope.

Discussion: DALL-E 2 refused to accept this caption and gave a warning that it violated policy, probably because it contains the word "war". We therefore rephrased it as example 5.A.

**Example 5.A:**

Caption: A donkey and an octopus are playing a game. The donkey is holding a rope on one end, the octopus is holding onto the other. The donkey holds the rope in its mouth. A cat is jumping over the rope.

Images:

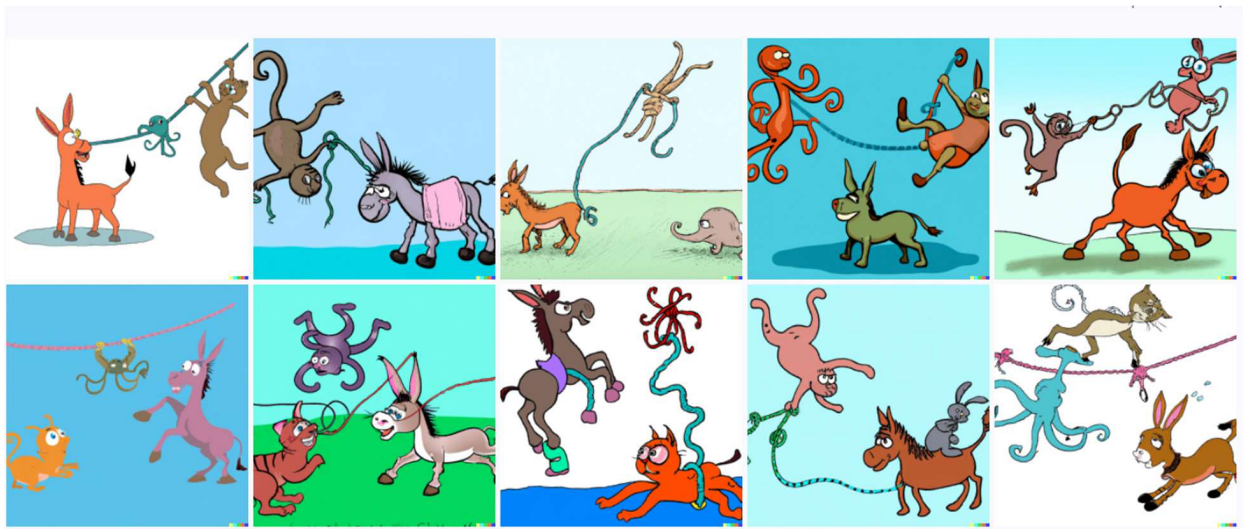

Discussion: All of the image contain all of the components of the caption – a donkey, a cat, a rope, and a creature that is multi-legged, though none of the images show an octopus with eight legs. None gets more than one of the stated relations right.

**Example 6:**

Caption: A pear cut into seven pieces arranged in a ring.

Images:

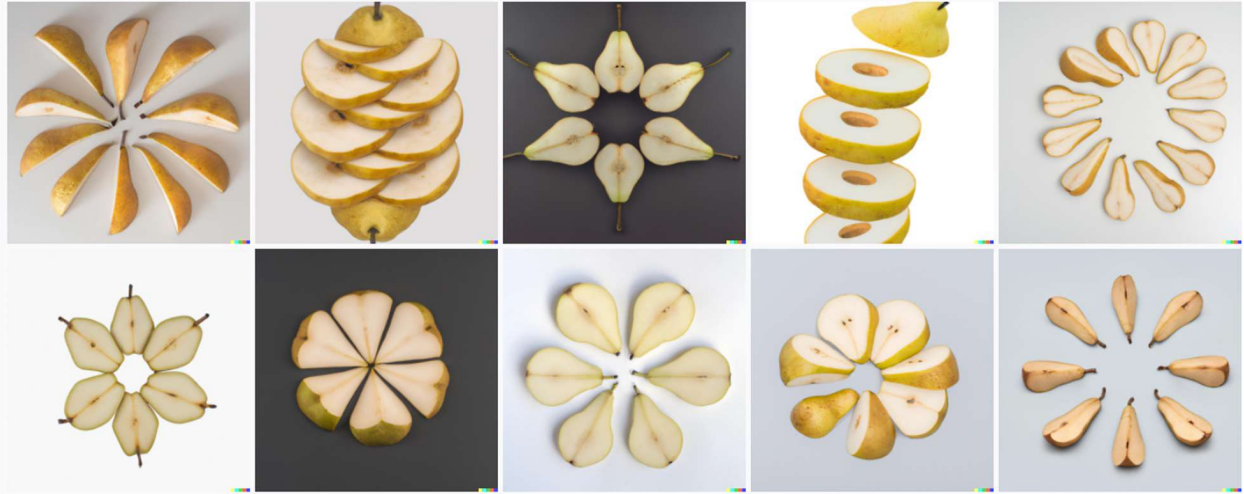

Discussion: All of the images show pieces of a pear and all but 2 and 4 show those pieces arranged as a ring. Image 9 successfully shows seven pieces, although it seems impossible for them to be pieces of a single pear. (By contrast, image 10 is a single pear cut into eight pieces.)

When we gave this caption to Google Images, all of the results returned showed a ring with a pear-cut diamond, and we thought that DALL-E 2 might make the same mistake, but it avoided that trap.

**Example 7:**

Caption: A tomato has been put on top of a pumpkin on a kitchen stool. There is a fork sticking into the pumpkin. The scene is viewed from above.

Images:

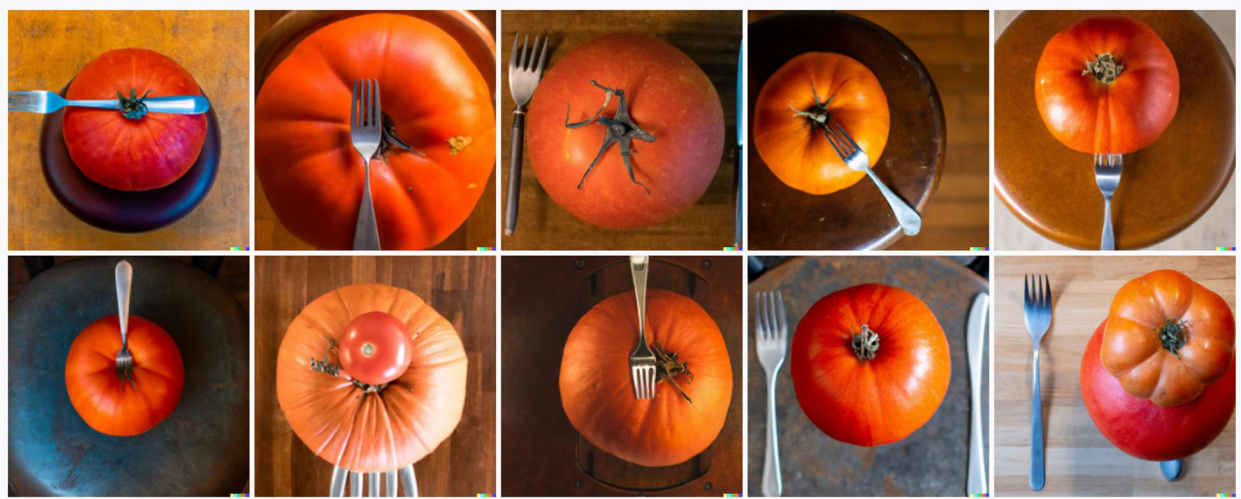

Discussion: DALL-E 2 got the viewpoint specification consistently which we had thought would be challenging. Only images 7 and 10 show both the tomato and the pumpkin, and only images 5, 6 and perhaps 7 show the fork stuck into the pumpkin.

**Example 8:**

Caption: An elephant is behind a tree. You can see the trunk on one side and the back legs on the other.

Images:

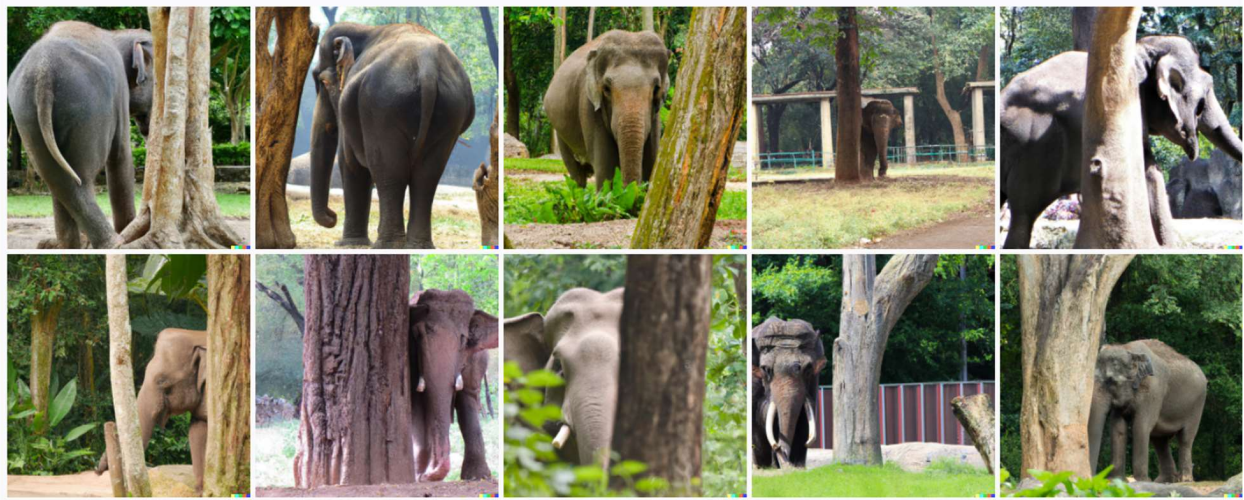

Discussion: This example was based on an actual photograph shown to the authors by Georgia Giokxari. All the images correctly show an elephant and a tree, and in images 3, 5, 6, 8, and 10, the elephant is behind the tree, or at least further from the viewer. Only image 5 shows it in the specified position.

**Example 9:**

Caption: In late afternoon in January in New England, a man stands in the shadow of a maple tree.

Images:

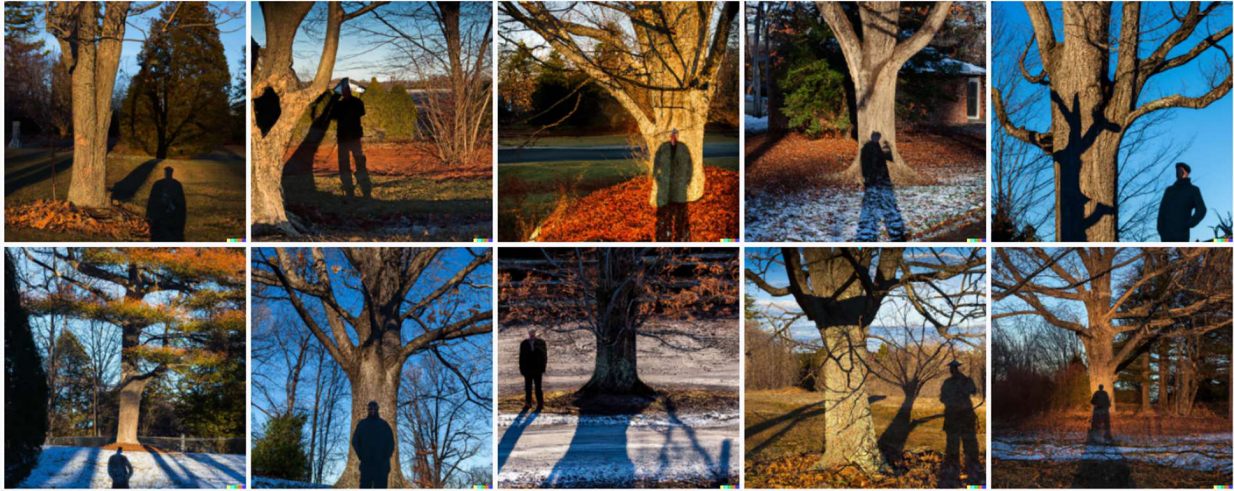

Discussion: DALL-E 2 correctly inferred that a maple tree in January has no leaves, which we had thought might be challenging. All the images show a man, and shadows, and, except in image 6, the shadows seem to align properly with one another and with the implicit lighting. However, in none is the man in the shadow of the tree, as specified. Moreover, image 3 seems unclear as to whether what is being shown is the man or his shadow; the coloring in the image in part reflects the man and in part reflects his background, as if he were translucent.

**Example 10:**

Caption: An old man is talking to his parents.

Images:

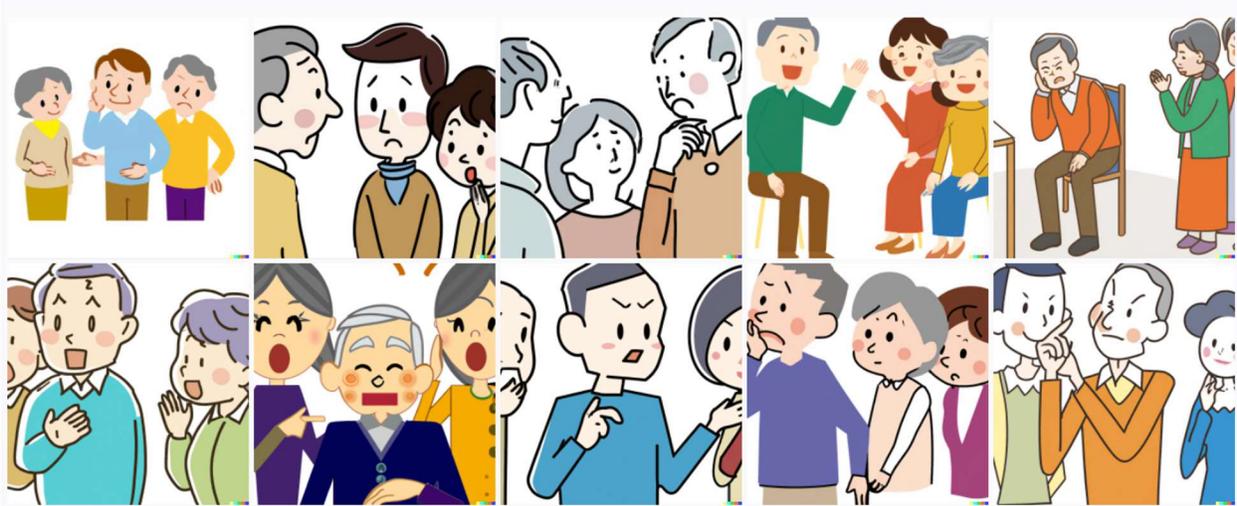

Discussion: In none of these images did DALL-E successfully infer that image should show an old man with two even older people (which was intended as a test of commonsense reasoning).

**Example 11:**

Caption: A grocery store refrigerator has pint cartons of milk on the top shelf, quart cartons on the middle shelf, and gallon plastic jugs on the bottom shelf.

Images:

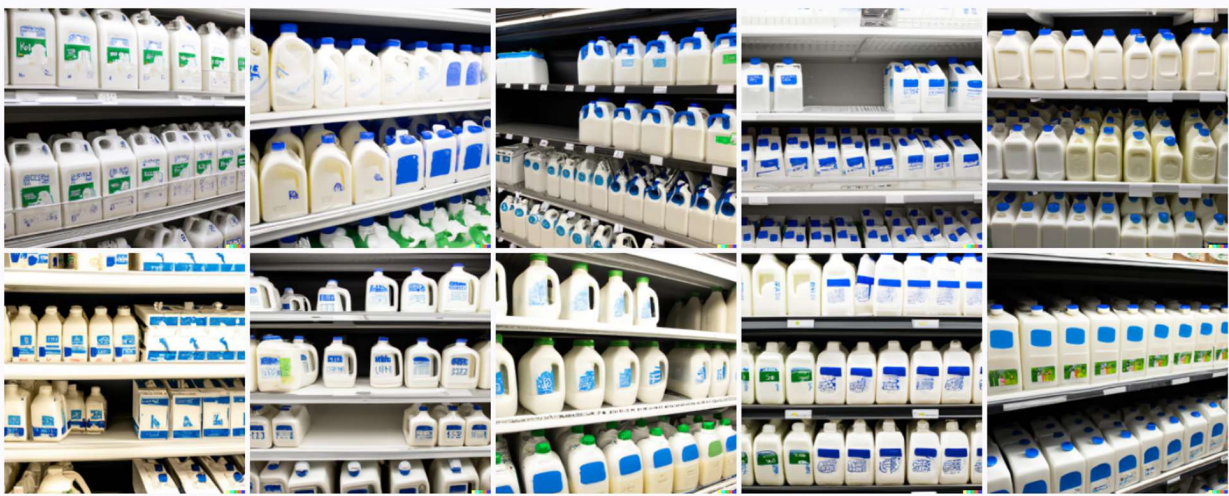

Discussion: The images all show a grocery store refrigerator, and most show three shelves, but none has the correct relative sizes.

**Example 12:**

Caption: A couple in formal evening wear going home get caught in a heavy downpour with no umbrellas

Discussion: DALL-E 2 chose to generate these in a photographic style, perhaps using images of actual people. OpenAI's policy prohibits publishing photographic images produced by DALL-E 2 that show people who might be recognizable. All ten images show a man and woman in formal evening wear in a downpour. In two of the images, the couple are underneath an umbrella but not in the remaining eight. In one image the couple seem to be indoors; this is a failure of commonsense physical reasoning., In one the streets are wet but it is not clear that it is still raining. However, six of the ten images unquestionably conform to the specification.

We ran the example a second time, specifying that the image should be an oil painting to avoid this problem.

**Example 12.A:**

Caption: An oil painting of a couple in formal evening wear going home get caught in a heavy downpour with no umbrellas.

Images:

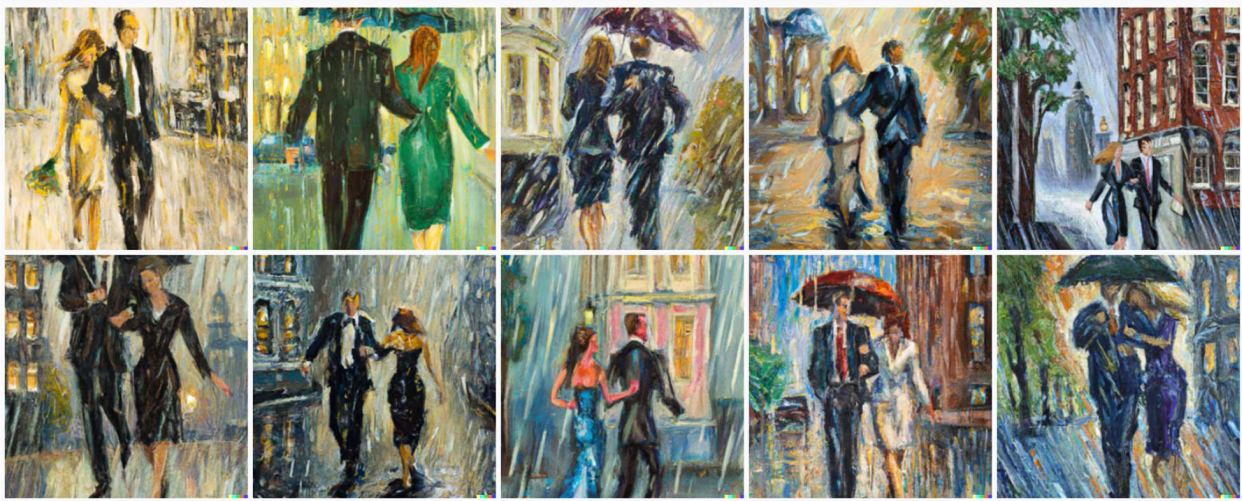

Discussion: Five of the ten images are exactly correct. In the other five, however, DALL-E2 has missed the specification of 'no umbrella.' " (We also ran a third attempt at this, with a caption that was probably identical and certainly very similar, but inadvertently not recorded. The outcome of that attempt was very similar in style; two of the images were correct but the other eight erroneously showed the couple with an umbrella.)

**Example 13:**

Caption: Paying for a quarter-sized pizza with a pizza-sized quarter.

Images:

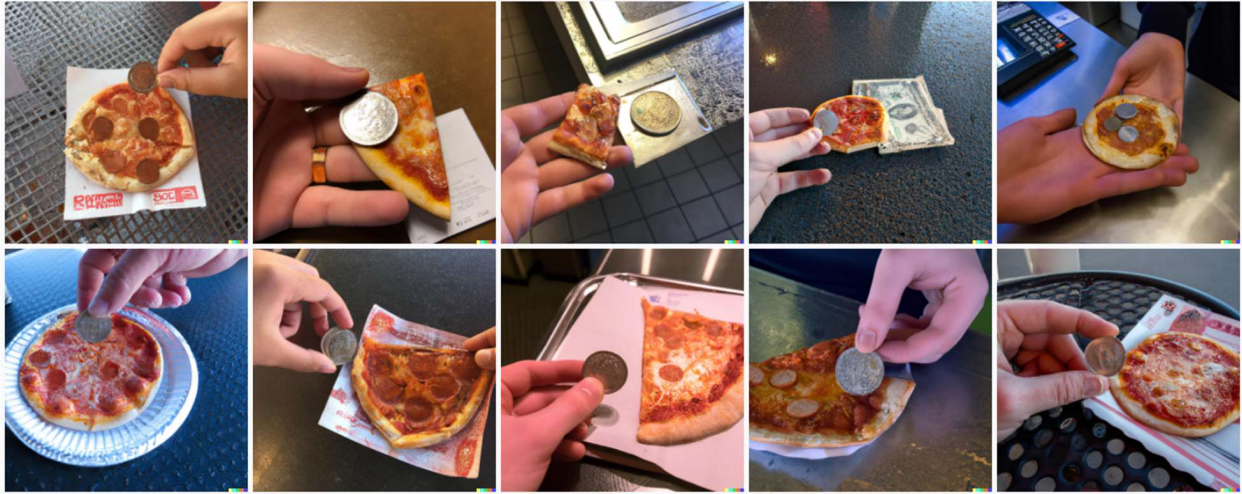

Discussion: This caption is taken from an example of Rips (1989). (There is a charming line drawing of it by Maayan Harel in (Marcus & Davis, 2019).) All the images here show a pizza and a quarter, and the pizzas are mostly quite small; though not certainly not the size of a quarter, perhaps they are a quarter of the size of a regular pizza, in terms of diameter. But none shows a pizza-sized quarter.

**Example 14:**

Caption: Wild turkeys in a garden seen from inside the house through a screen door.

Images:

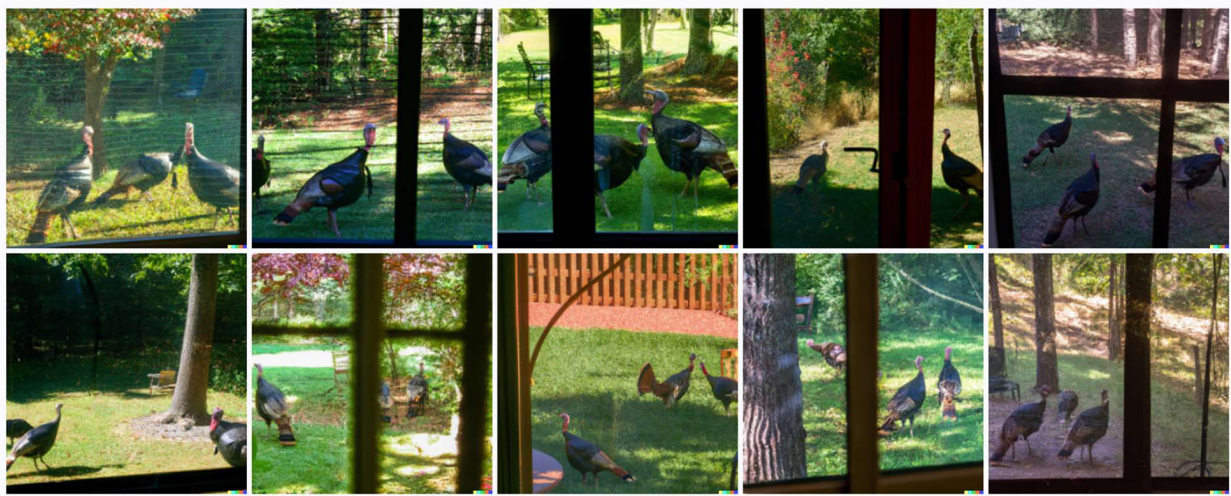

Discussion: The challenge here was to indicate that the turkeys were being viewed through a screen door. DALL-E2 succeeded in images 1 and 2 in indicating the screen or something similar. The remaining images looks as though the turkeys were being viewed through glass. None of the images clearly indicate that it is being viewed through a door rather than a window.

## Acknowledgements

Thanks to OpenAI for permission to use these images and to Aditya Ramesh for helpful clarifications. Scott Aaronson's research is supported by a Vannevar Bush Fellowship from the US Department of Defense, a Simons Investigator Award, and the Simons 'It from Qubit' collaboration.